\documentclass[10pt,twocolumn,letterpaper]{article}

\usepackage{wacv}              

\usepackage{graphicx}
\usepackage{amsmath}
\usepackage{amssymb}
\usepackage{booktabs}

\usepackage{float}
\usepackage{url}
\usepackage{rotating}
\usepackage{listings}
\usepackage[accsupp]{axessibility}

\usepackage[pagebackref,breaklinks,colorlinks]{hyperref}

\usepackage[capitalize]{cleveref}
\crefname{section}{Sec.}{Secs.}
\Crefname{section}{Section}{Sections}
\Crefname{table}{Table}{Tables}
\crefname{table}{Tab.}{Tabs.}

\def\wacvPaperID{*****} 
\def\confName{WACV}
\def\confYear{2024}

\begin{document}

\title{Image and AIS Data Fusion Technique \\ for Maritime Computer Vision Applications}

\author{Emre Gülsoylu\\
    University of Hamburg\\
    \and
    Paul Koch\\
    Fraunhofer CML\\
    \and
    Mert Yıldız\\
    Fraunhofer CML\\
    \and
    Manfred Constapel\\
    Fraunhofer CML\\
    \and
    André Peter Kelm\\
    University of Hamburg\\
    {\tt\small \{emre.guelsoylu, andre.kelm\}@uni-hamburg.de}\\
    {\tt\small \{paul.koch, mert.yildiz, manfred.constapel\}@cml.fraunhofer.de}
}

\maketitle

\begin{abstract}

    Deep learning object detection methods, like YOLOv5, are effective in identifying maritime vessels but often lack detailed information important for practical applications. In this paper, we addressed this problem by developing a technique that fuses Automatic Identification System (AIS) data with vessels detected in images to create datasets. This fusion enriches ship images with vessel-related data, such as type, size, speed, and direction. Our approach associates detected ships to their corresponding AIS messages by estimating distance and azimuth using a homography-based method suitable for both fixed and periodically panning cameras. This technique is useful for creating datasets for waterway traffic management, encounter detection, and surveillance. We introduce a novel dataset comprising of images taken in various weather conditions and their corresponding AIS messages. This dataset offers a stable baseline for refining vessel detection algorithms and trajectory prediction models. To assess our method's performance, we manually annotated a portion of this dataset. The results are showing an overall association accuracy of 74.76 \%, with the association accuracy for fixed cameras reaching 85.06 \%. This demonstrates the potential of our approach in creating datasets for vessel detection, pose estimation and auto-labelling pipelines.
   
\end{abstract}

\section{Introduction}
\label{sec:intro}
Maritime computer vision datasets are limited \cite{ward2018ship}, particularly due to costly manual annotation and lack of available images. 
Considering that the state-of-the-art Deep Learning (DL) object detection approaches like YOLOv5 \cite{yolov5} perform well for localizing maritime vessels, a technique to create maritime computer vision datasets can be developed. However, these object detection models can only provide limited maritime-related classes out-of-the-box. Applications like maritime autonomous surface ships (MAAS) require contextualized information such as the type, dimensions, speed, and course of a vessel \cite{fusion_camera_ais}. A system that matches AIS data with detected vessels from an object detection algorithm can help estimate the pose of the vessel, support in performing collision avoidance \cite{collision_avoidance} and aid in the creation of extensive maritime computer vision datasets.

In this paper, an easy to use technique that fuses Automatic Identification System (AIS) data with webcam-based images was developed. The technique adapts an already existing homography based coordinate transformation to associate detected ships in image space and available location information in world coodinates \cite{carrillo2022ship}. However, this paper extends the existing technique for rotating cameras. The proposed technique enables a scalable way to fuse images from fixed and panning cameras with the extensive amount of data provided by AIS. This type of system can support the creation of maritime computer vision datasets with less to no human intervention needed. Moreover, alterations on the technique have been made to ensure accurate spatial resolution when transforming from image-to-world coordinates.

A dataset has been created with this method and made publicly available \url{https://github.com/egulsoylu/image-ais-fusion}.


\section{Background}
\label{sec:background}

    \subsection{Automatic Identification System} 
    \label{sec:ais}
    The Automatic Identification System \cite{imo2001solas} is a radio-based communication system for the exchange of ship related parameters between ships and vessel traffic services (VTS). AIS was initially introduced to improve efficiency and safety, especially for collision avoidance in maritime navigation \cite{promises-perils}. Besides collision avoidance, it is currently used for various purposes, including maritime surveillance, \cite{fusion_camera_ais}, waterway management \cite{water_management}, and environment protection \cite{environment_protection}. As defined in the International Convention for the Safety of Life at Sea (SOLAS) \cite{imo2001solas}, the transmission of AIS messages is mandatory for certain vessels, including cargo ships, passenger vessels, and in some cases, fishing vessels. Since there are webcams based on the banks of the Elbe river that provide publicly available video streams, a huge volume of images can be collected and fused with AIS data.

    
    Although AIS messages have issues with reliability and manipulation, they still provide valuable data for maritime navigation. Some of the well-known issues of AIS are uncertainty associated with technical equipment for Global Positioning System (GPS) \cite{gps}, temporal differences \cite{promises-perils}, and malicious crew that switches off the transceiver \cite{sar_ship_detection_fusion}. 
    Moreover, data redundancy \cite{data_redundancy}, and noise \cite{ais_noise} are the problems mentioned by previous studies. The range and the equipment significantly impact GPS-related issues, which can be minimized by applying trajectory correction methods \cite{predicting_ship_trajectory, trajectory_prediction_waterway}. As AIS messages contain detailed information about the vessel, matching images with AIS data can benefit maritime informatics in training an object detection model to infer more information from an image and support marine traffic against the reliability issues of AIS.


    \subsection{Object Detection for Vessels}
    \label{sec:vessel_detection}
    Object detection models can effectively localize and classify vessels in various images. Adaptations of DL vessel detection methods for maritime applications include Gupta et al.'s integration of Support Vector Machine, bag of features, and CNNs \cite{gupta2021ship}, and Li et al.'s  lightweight model modifying YOLOv3 \cite{li2020lightweight}. Chang et al. enhanced YOLOv3 for better ship detection in both visible and infrared images \cite{chang2022modified}, and Xie et al. developed a more efficient network by integrating YOLOv4 with several advanced techniques, achieving high performance with significantly fewer parameters \cite{xie2022yolov4}.

    As can be seen, the vessel detection field is very active, with different methods being proposed on a regular basis, most of which depends on YOLO family object detection methods. In terms of vessel localization and maritime-related classification, these modifications improves the performance and sometimes efficiency. However, complex tasks require more than classifying a vessel type and this paper focuses on fusion of bounding boxes with AIS data. Therefore, a more recent version of a general purpose object detection model, YOLOv5 was fine-tuned for this paper.

    \subsection{Fusion of Images and AIS Data}
    The literature shows four directly related works on the fusion of camera-based images and AIS data. National Land Survey of Finland worked on sensor fusion for autonomous vessel navigation \cite{fusion_for_autonomous_2021}, funded by European Space Agency. However, this work's details are unclear due to the lack of published material. 
    
    Lu et al. \cite{fusion_camera_ais} introduced a framework for vessel identification by fusing images with AIS data. They proposed a distance estimation method based on the ship size by monocular vision. To improve the fusion accuracy, a method that is commonly used by the seaman called Dead Reckoning (DR) was utilised. It is a method to predict the current position based on a ship's previously observed speed, heading, and course. The most significant limitation of this work is the distance and bearing estimation, which leads to lower accuracy in matching the bounding boxes of vessels with related AIS messages. As the distance is predicted by the ship's overall length (LOA) and the width of the detected bounding box, the accuracy of distance estimation depends on the ship's heading and is highly sensitive to the localisation performance. Since the dimensions of bounding boxes can be affected by the background, water reflections, and weather conditions in general, distance estimation based on bounding boxes is not reliable. Moreover, this method produces incorrect results when the azimuth angle between the ship's heading and the camera is small. In other words, if a vessel is perpendicular to the camera's image plane, the width of the bounding box will represent the vessel's breadth instead of LOA.
    
    Qu et al. \cite{intelligent_fusion_camera_ais} introduced another framework for the fusion of camera-based vessel detection and AIS data. They utilise a YOLO-based vessel detection network and arrival time estimation to associate AIS messages with the detected vessels. The arrival time estimation takes the distance between the camera and a vessel into account for predicting the time the vessel will be in the field of view.

    Carillo-Perez et al. \cite{carrillo2022ship} propose a method for georeferencing ship masks that uses the homography method to transform coordinates from image coordinates to world coordinates  along with a novel dataset for maritime monitoring that contains images with taken by a static camera. They investigated instance segmentation methods focusing on more robust performance such as Mask-RCNN and DetectoRS as well as faster models such as YOLACT, Centermask-Lite. Their method of calculating the homography matrix relies on manually detecting antennas on each vessel in the image and matching them to the associated AIS message. The creation of pairs of image coordinates and world coordinates in this way is based on an assumption: The location information from the AIS data is reliable. However, as discussed in section \ref{sec:ais}, AIS data may not be reliable and may lead to errors in the estimation of the homography matrix. Also, this method would only work for fixed cameras, and more steps are required to apply it to a panning camera. 

    In this paper, we have generalised the method proposed by Carillo-Perez et al. \cite{carrillo2022ship} on the panning cameras and created a novel dataset using this method which is publicly available. Moreover, we created image space-world coordinates pairs for the calculation of homography matrix with artificial structures that does not move such as concrete levee crowns, pier poles and corners of buildings. The improvements on georeferencing method using homography is discussed in Section \ref{sec:methodology} and the details about the dataset is presented in Section \ref{sec:dataset}.

\section{Methodology}
\label{sec:methodology}
\begin{figure*}
        \centering
        \includegraphics[width=\linewidth]{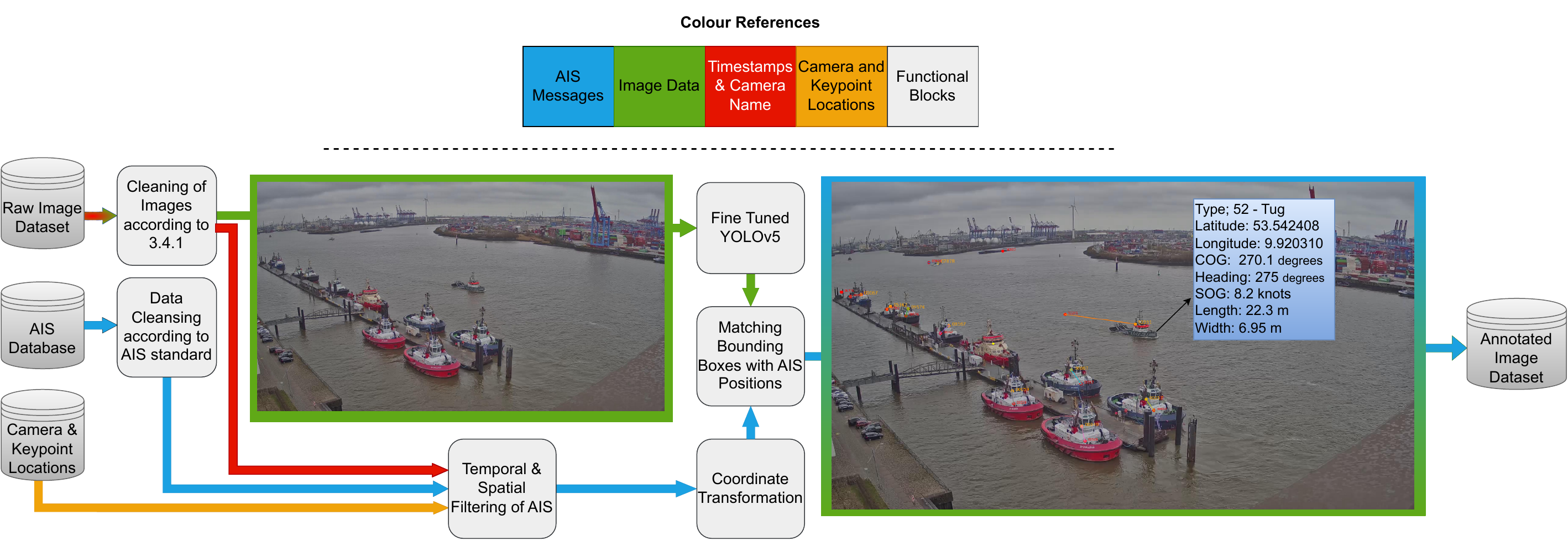}
        \caption{The pipeline for the technique. The technique takes an image, its timestamp and AIS messages of that day. After generating bounding boxes of the vessels on the image and filtering the AIS data, the system associates the predicted locations with the predicted bounding boxes. Red labels represent the location prediction based on AIS data and the unique identifier number. Orange labels represent the bounding box centre and associated unique identifier number.}
        \label{fig:system_overview}
    \end{figure*}

    The proposed procedure has two main steps, (1) vessel detection and (2) coordinate transformation to match vessel images with related AIS messages. First, the images are preprocessed to filter irrelevant images out. Amongst the filtered images, an image is fed into a fine-tuned YOLO model and bounding boxes are extracted. The image's timestamp is taken with the camera location, and AIS messages are filtered based on the time and location information. Then bounding boxes are matched with AIS messages, and the results are saved. The overview of the pipeline is shown in Figure \ref{fig:system_overview}.

    \subsection{Object Detection for Vessels}
        Object detection for vessels is one of the two main prerequisites for the fusion of images and AIS messages. As mentioned in Section \ref{sec:vessel_detection}, various YOLO models and their modified variants have been successfully used for vessel detection. Also, Lu et al. \cite{fusion_camera_ais} demonstrated that a fine-tuned YOLOv5 model performs well for maritime surveillance. In this paper, similar to Lu et al.'s work, a YOLOv5XL model was fine-tuned with the manually annotated dataset by changing the last layer from 80 neurons to only one, which represents the "Vessel" class. The actual vessel types are redundant as this information can be extracted from successfully matched AIS messages. Before fine-tuning, the mean average precision (mAP) of the model was 0.332 at an intersection over union (IoU) threshold of 0.5 and fine-tuned version performs 0.951 mAP at 0.5 IoU threshold on the test set. 
    
    \subsection{Transformation Between World Coordinates and Image Coordinates}
        AIS messages contain the spatial location of vessels according to World Geodetic System 1984 (WGS84) \cite{wgs_84}. Transforming the world coordinates (latitude, longitude) into image coordinates $ (x, y) $ enables the fusion of bounding boxes with AIS messages. In this section, two methods will be introduced for transformation between world coordinates and image coordinates. First, azimuth and distance estimation by interpolation will be presented in the Section \ref{sec:interpolation}. Then the homography approach will be explained in Section \ref{sec:homography} for different camera characteristics.
        
        \subsubsection{Keypoint Selection}
       
\begin{figure}
          \centering
          \includegraphics[width=\linewidth]{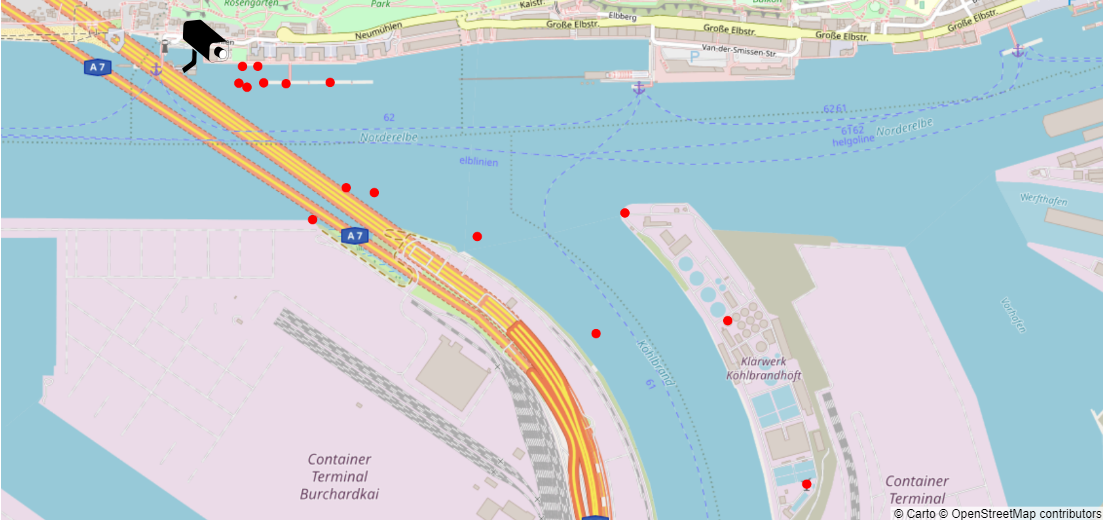}
           \includegraphics[width=\linewidth]{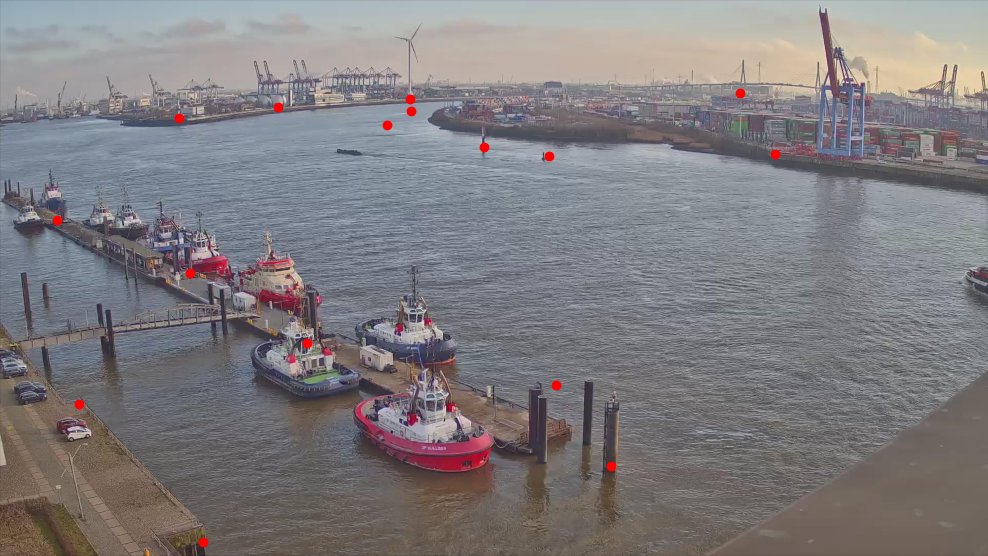}
          \caption{World (above) and image (below) coordinates for the keypoints of the Neumühlen camera represented with red points.}
          \label{fig:keypoints}
        \end{figure}

       For each camera, we selected an image with optimal visibility where the scene's features were distinctly clear, in order to manually select keypoints. More than ten keypoints, such as fixed buoys, antennas, dock walls, if applicable, significant landmarks (e.g., Elbphilharmonie), were selected for each image. Image coordinates of each key point were noted, and their corresponding world coordinates were detected with the help of Google Maps. The azimuth angle and the distance were derived from world coordinates for each key point by using inverse geodetic function with WGS84 ellipsoid \cite{wgs_84}. Figure \ref{fig:keypoints} shows the keypoints on the map and the image for the Neumühlen camera.
        
        \subsubsection{Azimuth and Distance Estimation by Interpolation}
        \label{sec:interpolation}
        Azimuth is the horizontal angle between the north and a point of interest in degrees. Every image aligned with the horizon has a linear relationship between the azimuth and the pixel coordinates on the $ X $ axis. In the dataset, the images are almost perfectly aligned with the horizon and demonstrate the linear relationship between the $ X $ axis and the azimuth degrees. Since the camera and the ship locations are roughly known, linear interpolation between the image's left and right edge can map each pixel on the $ X $ axis with an azimuth angle. The distance can be estimated with interpolation approach assuming that the keypoints cover the related parts of the image. 
        
        \subsubsection{Homography}
        \label{sec:homography}
        The lack of intrinsic and some extrinsic parameters prevented the estimation of the camera matrix, which would be helpful for transformation from world coordinates to image coordinates. Nevertheless, the relation between two images of the same planar surface can be found with homography. In this paper, instead of two images of the same planar surface, an image and a map were used of the same area to estimate the homography matrix. Therefore, the transformation between the map coordinates, $lon$, $lat$, and the image coordinates, $x$, $y$ in pixels can be performed as shown in Equation \ref{eq:homography}, where $w$ is the scale factor and $h_{i,j}$ denotes the elements of the homography matrix. The value 1 is appended to the image coordinates to obtain their homogeneous matrix to avoid dimensional mismatch.
        \begin{equation}
        \label{eq:homography}
            \begin{bmatrix}
                w \, lat \\
                w \, lon \\
                w
            \end{bmatrix} = 
            \begin{bmatrix}
                h_{11} & h_{12} & h_{13} \\
                h_{21} & h_{22} & h_{23} \\
                h_{31} & h_{32} & h_{33}
            \end{bmatrix}
            \begin{bmatrix}
                x \\
                y \\
                1
            \end{bmatrix}
        \end{equation}
        
        \paragraph{Transformation in Fixed Cameras}
        First, the keypoints were detected manually to estimate the homography matrix for the fixed cameras. Then the corresponding world coordinates were detected from Google Maps. After having two sets of points in different coordinate systems, the homography matrix was estimated.
        Therefore, the world coordinates provided by AIS messages can be transformed into image coordinates which can be later matched with related bounding box centres.
        
        \paragraph{Transformation in Panning Cameras}
        Due to irregular image collection intervals, the images collected from panning cameras require localisation first. Since the images from various angles are present, a panorama for each panning camera was created by stitch images. After having a panorama image, the method followed for fixed cameras can be applied for homography estimation to transform world coordinates into panorama image coordinates. Unlike fixed cameras, there is one more step to take: transforming query image coordinates to panorama image coordinates. Each image was used as a query image and localised in the related panorama image with template matching, using cross-correlation. The transformation from query to panorama image coordinates was applied using the highest cross-correlation value as the offset. Thus the ships were localised in the query image based on the AIS message in two steps (1) world coordinates to panorama image coordinates, (2) query image coordinates to panorama image coordinates.

        Cross-correlation is sensitive to small changes as it is a pixel-level comparison \cite{debella2011sub}. To improve the localisation of query images in panoramas, a future extraction algorithm, Oriented FAST and rotated BRIEF (ORB) \cite{rublee2011orb}, was tried. However, it did not outperform the template matching approach. The main reason for this is that the images contain cranes, wind turbines and other structurally identical and moving objects, which limits the performance of the algorithm during the keypoint matching process. 
        
    \subsection{Bounding Box-AIS Message Association}
        \label{sec:association}
\begin{figure*}
          \centering
          \includegraphics[width=\linewidth]{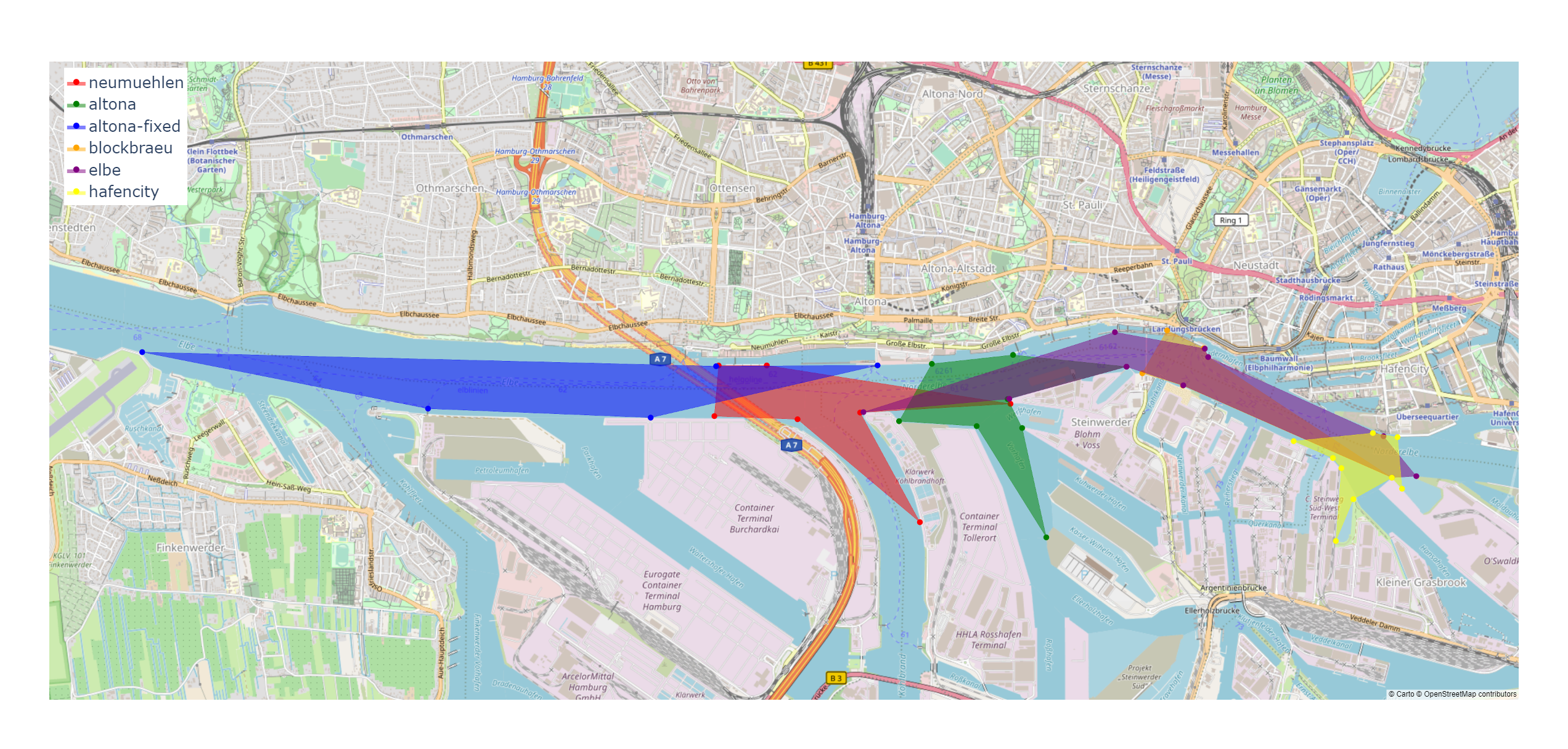}
          \caption{Regions of interest for each camera.}
          \label{fig:camera_roi}
        \end{figure*}

        AIS data is filtered by a region of interest defined manually for each camera and 30 seconds period before and after the image's timestamp. The region of interest is the whole area that panning cameras cover. Figure \ref{fig:camera_roi} illustrates the area for each webcam. Then filtered messages are compared with the bounding boxes in the related image. In this dataset, the images are less frequent than the AIS messages, and there is no one-to-one relation between the images and the messages. Therefore, the location of vessels is predicted by interpolation according to the image's timestamp, assuming that the vessel speed is constant since the time frame is not big enough to significantly change the speed. If there are not enough datapoints to interpolate for a vessel at a time, then the location transmitted in AIS message is used. Applying interpolation also helps reducing the negative effects of the unstable video stream problem. After getting the predictions, AIS messages are assigned to the closest bounding box in the image by using nearest neighbour search based on k-dimensional tree \cite{bentley1975multidimensional}.
        
        
    \subsection{Dataset}
    \label{sec:dataset}
        The visible light images in this dataset were collected from five public webcam streams between 11.01.2022 and 20.03.2022 during the daylight hours.
        These webcams are located on the bank of Elbe river and its tributary Norderelbe, near the centre of Hamburg. Three of these webcams are fixed, while the other three are panning, one of which demonstrates both characteristics. 
        Fixed cameras are those that do not change direction and always point in one direction. Although the wind changes their direction slightly from time to time, these changes are so small that they can be ignored. Panning cameras, on the other hand, change their direction with yaw rotation and therefore cover a wider area than the fixed cameras. 
        
        One of the cameras shows two characteristics: It is primarily a panning camera, however, after a short transition, it focuses on a building for a minute and behaves like a fixed camera. Therefore, this camera can be treated as two cameras with different characteristics. Although the camera angle is slightly different because of the panning movement during the transition from panning to fixed characteristic, it is still closer to a fixed camera than a panning camera. The images produced by this camera were divided into three classes with appearance-based classification 
        (1) Panning images, which were treated as query images and compared with a panorama, (2) Fixed images, which contain the images of Dockland Office, (3) Transition images, which were captured during the transition.

        
        
        The histogram of each image is compared with other histograms that belong to one of these classes. The other histograms are from a subset of the images that were captured on 11.01.2022. In this subset, 25 panning, 24 fixed, and 17 transition image histograms are present. A class was assigned based on the smallest Euclidean Distance between the histograms. Before this method, Structural Similarity Index (SSIM) \cite{wang2004ssim} and edge difference zero-count were used, but these methods did not provide satisfactory results. Although the edge detection method seemed logical initially, as the transition images were out of focus, it could not outperform the histogram comparison.
        Table \ref{tab:camera_characteristics} shows the characteristics of the webcams.
    
        \begin{table*}
        	\centering
        	\begin{tabular}{|l|l|r|r|r|}
        		\hline
        		\textbf{Name} & \textbf{Type} & \textbf{Resolution (px)} & \textbf{Location (Lat, Lon)} & \textbf{Direction} \\\hline\hline
        		Altona & Fixed & 1280 x 720 & 53.54387, 9.94275 & South West \\\hline
        		Blockbräu & Fixed & 1280 x 720 & 53.54553, 9.96957 & South East \\\hline
        		Neumühlen & Fixed & 1920 x 1080 & 53.54388, 9.91692 & South East \\\hline \hline
        		
        		Altona & Panning & 1280 x 720 & 53.54387, 9.94275 & South East-South \\\hline
        		Elbe & Panning & 1280 x 720 & 53.54722, 9.96338 & South East-South West \\\hline
        		Hafencity & Panning & 1920 x 1080 & 53.53903, 9.99345 & South-South West \\\hline
        	\end{tabular}	
    	\caption{Characteristics of the webcams on the banks of the Elbe river.}
    	\label{tab:camera_characteristics}
        \end{table*}
    
        There are a few drawbacks to collect images from a video stream. First, network problems affect image collection significantly, as a delay can occur both on the server and the client sides. Delays affect the timestamps during image collection, which result in predictions that are significantly in front of the vessel. Moreover, some images were repeated without a clear repetition pattern due to an unreliable data stream. Because of the random nature of image repetition, a preprocessing step was used for filtering them out. The images were compared with 20 previous frames by using histogram comparison, and the duplicate images were removed. As the traffic is not always dense, the images did not always have a ship in them. By using both the fine-tuned YOLO model and checking the presence of AIS messages for each camera's RoI, the images without any ships were detected and removed. These preprocessing steps resulted in the removal of images with rain drops which covers the whole lens, dark lighting conditions and heavy fog.
        
        Panning cameras got affected the most by the infrequent data collection. Since the images were collected in various intervals, the camera angle was different each time an image was captured, making it harder to handle images from panning cameras individually. Panoramas were created for each camera, and each individual image was compared with the corresponding panorama image to overcome this problem (see Section \ref{sec:homography}).
        



        
        Due to the difficulty of annotating the high amount of collected images, only a subset from collected images were annotated in YOLO format with \nolinkurl{"makesense.ai"}. In total, 1515 images were annotated of which 1062 images were used during training, 299 images were during validation, and 154 images during test while fine-tuning the YOLOv5 model. After bounding box annotation, the unique identifier numbers were assigned for bounding boxes manually with the help of AIS data provided. During the annotation process, \nolinkurl{"marinetraffic.com"} was also used as it provides images and MMSI numbers of vessels. It is worth mentioning that the colour changes of the ferries due to advertisements made the annotation process more difficult. 1658 bounding box-AIS message pairs were created in 381 images.

        Besides image\_id, bbox, category\_id and unique\_id the annotations contain details about the vessel taken from the associated AIS message. A sample annotation is shown in the Listing \ref{annotation_snippet}.

        \begin{lstlisting}[label=annotation_snippet, caption=A sample annotation snippet for the vessel information that was extracted from the associated AIS message.]
        "annotations": [
        {
            "image_id": 0,
            "bbox": [
                363.0,
                602.0,
                199.0,
                56.0
            ],
            "vessel_info": {
                "type": 70,
                "latitude": "53.542968",
                "longitude": "9.935401",
                "heading": 268.3,
                "course": 271.38,
                "length": 29,
                "width": 7,
                "speed": 8.73,                
            }
        },
        ...
        ]
        \end{lstlisting}

\section{Results and Discussion}
    
    \subsection{Coordinate Transformation}
    Preliminary results show that homography outperforms azimuth and distance estimation by interpolation, as interpolation needs full coverage of the related parts of the image. However, it is hard to find keypoints that cover the whole area. Extrapolation can be a solution for this in exchange of precision. Because of this reason, homography is used as the technique for transforming coordinates.

    \begin{table*}
    	\centering
    	\begin{tabular}{|p{19mm}|p{17mm}|p{20mm}|p{20mm}|p{20mm}|p{19mm}|p{18mm}|}
    		\hline
    		\textbf{Camera Name} & \textbf{Camera Type} & \textbf{Maximum \newline Error (px)} & \textbf{Minimum Error (px)} & \textbf{Mean \newline Error (px)} & \textbf{Standard Deviation} & \textbf{Keypoint Count}\\\hline\hline
    		Blockbräu & Fixed & 161.51 & 4.24 & 39.97 & 31.84 & 27 \\\hline
    		Altona & Fixed & 40.26 & 4.12 & 16.58 & 9.46 & 15 \\\hline
    		Neumühlen & Fixed & 31.38 & 0.00 & 11.39 & 10.29 & 16 \\\hline \hline
    		Elbe & Panning & 451.46 & 43.46 & 146.85 & 123.38 & 11 \\\hline
    		Altona & Panning & 267.32 & 7.28 & 90.04 & 88.29 & 16 \\\hline
    		Hafencity & Panning & 245.00 & 17.80 & 124.04 & 63.02 & 17 \\\hline
    	\end{tabular}	
	\caption{Coordinate transformation error from world coordinates to (panorama) image coordinates.}
	\label{tab:coordinate_transformation}
    \end{table*}
    
    Table \ref{tab:coordinate_transformation} shows the coordinate transformation error introduced by homography. As the table shows, the fixed cameras produce less error compared to panning cameras because of the distortion introduced during the panorama creation in the coordinate transformation step. Also, considering that the panorama image is significantly bigger than an individual image, they are more likely to have bigger errors. It is clear that an increase in the number of keypoints affects the performance positively. It is to be noted that continuous values for the pixel errors stem from the homography matrix and are purely used for evaluation purposes.

           
    \subsection{Image-AIS Matching}
        The rate of successfully associated bounding box-AIS message pairs was used to evaluate the performance of image-AIS fusion. The accuracy values were calculated as shown in equation \ref{eq:accuracy}. The Maritime Mobile Service Identity (MMSI) number assigned to the bounding box was manually verified using the vessel images at \nolinkurl{"marinetraffic.com"}. If the visual inspection can confirm that the detected vessel is associated with the correct MMSI, the association is counted as correct. For the privacy reasons, the public dataset does not contain MMSI but a unique identifier number.
        
        As the panorama images have bigger regions of interest, AIS messages that do not land on the query image are not filtered out. Since AIS messages are sequentially assigned with the closest bounding box, sometimes AIS messages from outside the query image can be associated with the closest bounding box, which is usually an incorrect match. Vessels that transmit at large intervals or not at all, for example moored vessels, create a problem as their bounding box can be detected but cannot be associated with an AIS message correctly.
        \begin{equation}
        \label{eq:accuracy}
            Accuracy = \frac{Correct \, Associations}{Total \, Number \, of \, Pairs} * 100
       \end{equation}

        \begin{table}
    	\centering
        	\begin{tabular}{|p{16mm}|p{8mm}|p{13mm}|p{13mm}|p{13mm}|}
        		\hline
        		\textbf{Camera Name} & \textbf{Image Count} & \textbf{Accuracy (\%)} & \textbf{Correct \newline Pred.} & \textbf{Total \newline Pred.} \\\hline\hline
        		Altona & 60 & 34.52 & 29 & 84 \\\hline
        		Altona-F & 61 & 43.00 & 43 & 100 \\\hline
        		Blockbraeu & 70 & 77.25 & 180 & 233 \\\hline
        		Elbe & 60 & 24.48 & 24 & 143 \\\hline
        		Hafencity & 60 & 80.33 & 49 & 61 \\\hline
        		Neumühlen & 70 & 94.13 & 625 & 664 \\\hline \hline
        		\textbf{Total} & \textbf{381} & \textbf{74.79} & \textbf{961} & \textbf{1285} \\\hline
        	\end{tabular}	
    	\caption{The accuracy of successful associations for cameras. Altona-F is the fixed behaviour of the Altona camera.}
    	\label{tab:association_accuracy}
        \end{table}
    
    Table \ref{tab:association_accuracy} shows the accuracy of successful associations for each camera. The overall performance of the system is 74.79 \% accuracy in matching bounding boxes with AIS messages. For fixed and panning cameras, the system achieves 85.06 \% and 39.24 \%, respectively. 
    
    Fixed cameras outperform panning cameras because of the extra step of localisation of query images in panorama images, which fixed cameras do not have. Any bounding box-AIS message pairs that are present on an unsuccessfully localised query image are counted as unsuccessful matches whether they are successfully matched or not, as the localisation step comes prior to the matching step. 
    
    Tidal changes slightly affect the accuracy negatively since the keypoints were selected on sea level and the plane predicted by homography does not update itself accordingly. Weather conditions, such as fog and rain, can blur or entirely obstruct vessels in images. Although it is a rare scenario, sometimes birds can block the view of cameras. 
    
    The camera Elbe, which covers the widest area compared to the other cameras, is the worst performing one with an accuracy of 24.48 \%. This poor performance can be explained in two aspects. Considering the hardware aspect, Elbe is a panning camera located furthest from the river without a cover for the rain. It has the lowest number of keypoints, which results in the poor transformation from AIS message coordinates to image coordinates. Moreover, it introduces the highest amount of distortion when the panorama is created because of the wide area it covers, which results in poor performance for localising a query image in the panorama image. The accuracy of correct localisation of a query image is 58.33 \%. Considering these drawbacks, the Elbe camera requires more manual work for the selection of new keypoints, and a better technique of localisation of query images in panorama images.
    
    Neumühlen is a fixed camera that faces towards a pier where service boats and tug boats moor. More importantly, the direction of the camera is not perpendicular to the river. This makes the association easier as the error introduced by the unstable video stream becomes more tolerable because of the smaller displacement in the image, as mentioned in Section \ref{sec:dataset}. When the vessels move slower or moor, it is easier to correctly associate bounding boxes with AIS messages as long as they keep transmitting AIS messages which tug boats mostly do since they only moor for a short period of time. The slow but consistent traffic can explain the high performance of the Neumühlen camera.
    
    Comparing the results with Lu et al.'s framework, which obtains 69.35 \% without the short-time trajectory prediction method (DR), and 75.70 \% with DR, Qu et al.'s framework outperform Lu's framework with 81.42 \% overall accuracy. Our approach, on the other hand, outperforms both works for static cameras while the overall accuracy is slightly lower than Qu et al.'s framework. These works do not use a dataset consisting of images taken from web cameras with different and unknown parameters. Additionally, the datasets do not contain weather conditions such as fog and rain. Lu's dataset consists of images of the English Channel, while Qu's dataset consists of images of the Yangtze River, both of which are significantly broader than the Elbe river. Therefore, the traffic of the Elbe River is expected to be denser which requires more precision in the predictions and handling for the occluded vessels.

\section{Conclusion}
    In this paper, a technique was developed for creating maritime computer vision datasets by fusing ship bounding boxes with related AIS messages. The technique extends Carillo's technique \cite{carrillo2022ship} by adapting the homography-based coordinate transformation for non-static cameras, thus increases the number of application use cases. The technique consists of a fine-tuned YOLOv5 vessel detection model and a homography-based coordinate transformation module that can project world coordinates extracted from AIS messages onto images to match detected vessels' bounding boxes with AIS messages. 
    
    The results show that the proposed system achieves an accuracy of 85.06 \% with fixed cameras while achieving an overall accuracy of 74.79 \%. These results were achieved with images taken in various weather conditions during the day. Compared to the similar frameworks, it performs better for fixed cameras while can work with the panning cameras.

    Using the proposed technique, a dataset consisting of images and associated AIS messages can be created covering the Elbe River in Hamburg, a major hub and the third busiest port in Europe. Therefore, images offer a rich variety of ship types and multi-ship-encounter scenarios. Consequently, it stands as a valuable cornerstone for advancing maritime research through machine learning and artificial intelligence, serving as an indispensable resource for ongoing projects such as LEAS \cite{leas}, AUTOSHIP \cite{bolbot2020paving}, and AEGIS \cite{aegis}. Some possible use cases include inter alia research on elevating the precision of multi-sensor fusion techniques, thereby significantly enhancing navigational safety in the maritime domain and the creation of digital twins, opening doors to a multitude of possibilities for maritime system modeling and simulation, ultimately fostering a comprehensive understanding of the maritime ecosystem.

    \section{Future Work}

    The proposed dataset is particularly useful for designing auto-labelling pipelines to enhance existing object detection models or to solve pose estimation tasks. Current investigations use the annotated data and the homography matrix to generate labelled bounding box annotations in 3D world space, which are fed into keypoint detection networks to learn pose estimation from image space without the need to generate a homography matrix. 
    
    In terms of improving the proposed technique, the localisation of query images in panoramic images using template matching can be considered as the main limitation of this work. Improving this aspect of the fusion process of panning cameras would increase the accuracy significantly as the results show that the main problem is not the coordinate transformation but the template matching. This problem can be overcome by implementing feature-based localisation techniques for most cases. However, ships that occlude the landscape, or weather conditions such as fog and rain require a completely different approach. 
    
    
    Currently, keypoint selection for homography estimation is a process that has to be done manually. Considering that the dataset only contains images of the Elbe river and the camera locations are roughly known, it is possible to apply shape alignment methods to align the landmarks with the map with a similar approach of Shi et al. \cite{shape_alignment}. Although sports fields have proper shapes since they are human-made, in the future, trying to register the sea charts with images of Elbe can be fruitful. Moreover, ground-to-satellite image matching methods \cite{shi2022accurate} can be an area to explore for automatic keypoint selection. 

{\small
\bibliographystyle{ieee_fullname}
\bibliography{paper}
}

\end{document}